\documentclass[conference]{IEEEtran}
\IEEEoverridecommandlockouts
\def\BibTeX{{\rm B\kern-.05em{\sc i\kern-.025em b}\kern-.08em
    T\kern-.1667em\lower.7ex\hbox{E}\kern-.125emX}}

\usepackage{graphicx}
\usepackage{algorithm}
\usepackage[noend]{algorithmic}
\usepackage{array}

\usepackage{adjustbox}

\begin{document}

\title{The Canonical Interval Forest (CIF)  Classifier for Time Series Classification
\thanks{This work is supported by the UK Engineering and Physical Sciences Research Council (EPSRC) iCASE award T206188 sponsored by British Telecom.
}
}

\author{
\IEEEauthorblockN{Matthew Middlehurst}
\IEEEauthorblockA{\textit{School of Computing Sciences} \\
\textit{University of East Anglia}\\
Norwich, United Kingdom}
\and
\IEEEauthorblockN{James Large}
\IEEEauthorblockA{\textit{School of Computing Sciences} \\
\textit{University of East Anglia}\\
Norwich, United Kingdom}
\and
\IEEEauthorblockN{Anthony Bagnall}
\IEEEauthorblockA{\textit{School of Computing Sciences} \\
\textit{University of East Anglia}\\
Norwich, United Kingdom}
}


\maketitle

\begin{abstract}
  Time series classification (TSC) is home to a number of algorithm groups that utilise different kinds of discriminatory patterns. One of these groups describes classifiers that predict using phase dependant intervals.
  The time series forest (TSF) classifier is one of the most well known interval methods, and has demonstrated strong performance as well as relative speed in training and predictions. However, recent advances in other approaches have left TSF behind.
  TSF originally summarises intervals using three simple summary statistics.
  The `catch22' feature set of 22 time series features was recently proposed to aid time series analysis through a concise set of diverse and informative descriptive characteristics.
  We propose combining TSF and catch22 to form a new classifier, the Canonical Interval Forest (CIF). 
  We outline additional enhancements to the training procedure, and extend the classifier to include multivariate classification capabilities. 
  We demonstrate a large and significant improvement in accuracy over both TSF and catch22, and show it to be on par with top performers from other algorithmic classes.
  By upgrading the interval-based component from TSF to CIF, we also demonstrate a significant improvement in the hierarchical vote collective of transformation-based ensembles (HIVE-COTE) that combines different time series representations.
  HIVE-COTE using CIF is significantly more accurate on the UCR archive than any other classifier we are aware of and represents a new state of the art for TSC.
\end{abstract}


\section{Introduction}

    Time series classification (TSC) has experienced a rapid algorithmic advancement in predictive performance over the last decade.
    Prior to this, the one nearest neighbour with dynamic time warping distance classifier, with warping window set through cross validation (DTWCV), was considered the gold standard benchmark and difficult to beat. 
    However, a wide scale survey and evaluation of TSC literature~\cite{bagnall2017great}  found a number of classifiers able to significantly improve on DTWCV, and formed a taxonomy of similar approaches based on representations of discriminatory features. This taxonomy grouped algorithms into categories  based on the representations: distance-based; shapelet-based; dictionary-based; frequency-based; and interval-based (more details are provided in Section~\ref{sec:background}).  
    
    Three algorithms proposed subsequent to this evaluation have significantly higher accuracy than all of those evaluated in~\cite{bagnall2017great} and can claim to be among the state of the art for general-purpose TSC. These algorithms also represent the best of three alternative approaches to leveraging different representations for TSC. The hierarchical vote collective of transformation-based ensembles (HIVE-COTE)~\cite{lines2018time} encapsulates representations within modules containing algorithm(s) based on a single representation, then ensembles over these distinct modules. The time series combination of heterogeneous and integrated embeddings forest (TS-CHIEF)~\cite{shifaz2020ts} embeds multiple representations within the nodes of a decision tree, then ensembles many trees in a way similar to random forest. Thirdly, a bespoke representation for the dataset can be learned through deep learning approaches which have gained in popularity recently within TSC research.  InceptionTime~\cite{fawaz2019inceptiontime}, an ensemble of randomly initialised ResNet-style networks with Inception modules, is currently the strongest of these.

\begin{figure}
        	\centering
            \includegraphics[width=0.98\linewidth]{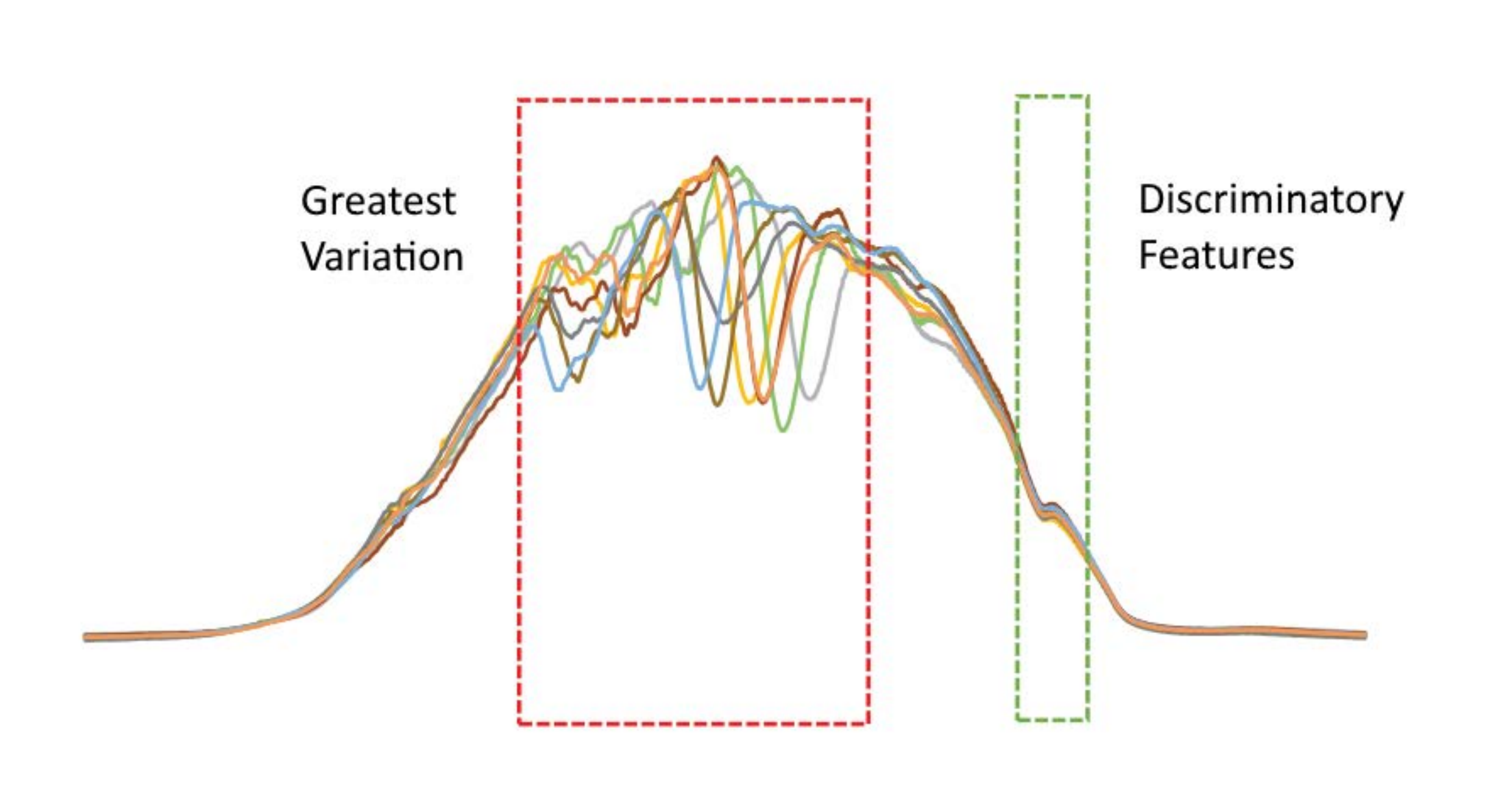}
            \caption{Time series from the EthanolLevel dataset. While the middle portion (visible light wavelengths) contains the greatest variation, a particular interval (into the near infrared) contains the most discriminatory information for ethanol concentration. An interval-based classifier may be able to compensate for the confounding effect of the visible spectrum. }
            \label{fig:ethanolExample}
        \end{figure}

    Our aim is to advance state of the art in TSC by improving one component of HIVE-COTE. HIVE-COTE consists of five modules, each of which represents the best in class (at the time of  publication in 2017) for the five different representations proposed in~\cite{bagnall2017great}. Our focus is on interval-based classifiers.  This class of algorithms derive features of series from randomly selected, but constant over series, intervals. Through taking multiple intervals and randomisation, interval-based classifiers are able to mitigate against regions of series that may confound other algorithms due to, for example, high noise or constancy. 
      These classifiers perform best when the particular location in time of discriminatory features is important for class membership.
    An example of this is shown in the EthanolLevel problem presented in Figure~\ref{fig:ethanolExample}. 
    This describes the visible-near infrared spectra of suspect spirits, where in this case the aim is to predict the concentration of ethanol in the sample. 
    Based on the resonances of ethanol around these ranges, a human expert would know to look at a particular interval of wavelengths. 
    While the noise in the center would confound whole series methods such as DTWCV, interval methods can better extract and summarise the discriminatory subseries.
    If we were to label the samples according to other factors, such as the presence of artificial caramel colouring, a different interval may be appropriate to use. 
    
    Time series forest (TSF)~\cite{deng2013time} was an early example of an interval-based forest of trees ensemble for TSC.
    TSF has been shown to be significantly better than DTWCV~\cite{bagnall2017great}, and represents the interval-based group of algorithms in HIVE-COTE.
    It has a number of secondary desirable properties it holds common with random forests: TSF is very fast to train;
    it is robust to parameter settings, with parameter tuning having no significant overall effect; and the trained forest can be post-processed to extract the importance of different intervals towards classification~\cite{deng2013time} - a useful tool for interpreting the model. However, TSF is, on average, the weakest classifier in the HIVE-COTE ensemble in terms of predictive power.  
    
    TSF works on only three basic summary stats for each interval: the mean; standard deviation; and slope. Recently, a set of 22 features, the canonical time-series characteristics (catch22)~\cite{lubba2019Catch22}, have been  proposed as good primitive summary measures for time series.
    These features cover a diverse range of concepts and are relatively quick to process. However, by themselves they are not particularly useful for classification when applied as a transform to the whole series. The resulting classifier with a random forest of 500 trees built on the transform is not significantly better than DTWCV. 
    
    Our primary contribution is to describe an algorithm for embedding catch22 features in an adaptation of TSF. This new classifier, the Canonical Interval Forest (CIF) is significantly more accurate than TSF; it is not significantly worse than the best recently proposed algorithms based on a single representation; and replacing TSF with CIF in HIVE-COTE results in significant improvement. We present results for HIVE-COTE with CIF that are significantly better than the best known alternatives for univariate classification, TS-CHIEF and InceptionTime.
    We additionally expand CIF for multivariate classification, and demonstrate CIF's strong performance in this relatively immature domain.
   
   Our secondary contribution, which facilitates the first, is to provide an evaluation and comparison of eight TSC algorithms described in Section~\ref{curr_sota}, using the recently expanded UCR archive~\cite{dau2019ucr}. These algorithms have been proposed since the bake off ~\cite{bagnall2017great}. All the code to do this is open source and all the code and results are available on the accompanying website. 
   
   
 

 
    The rest of this paper is structured as follows. 
    Section~\ref{sec:background} provides a technical background to TSC and related algorithms.
    Section~\ref{sec:cif} explains the changes made to TSF in order to include catch22 features and form CIF, and Section~\ref{sec:methodology} describes our experimental methodology.
    Section~\ref{sec:results} describes the results of investigations into the improvements made both by CIF over TSF, and to HIVE-COTE with CIF's inclusion over TSF. 
    In Section~\ref{sec:efficiency} we compare  the time and space complexity of a range of classifiers and describe how CIF can be used to aid interpretability and practical usage. We demonstrate the utility of CIF with a case study using data not currently in the UCR archive in Section~\ref{sec:usecase}. 
    Finally, in Section~\ref{conclusion} we conclude and discuss future work.
    
\section{Time Series Classification}
\label{sec:background}

    A time series is an ordered vector of continuous values that is typically taken over fixed intervals of time, but may be representative of other ordered data, e.g. an image outline. 
    An instance $(\textbf{X}, y)$ is a collection of $d$ time series $x$ with measurements at $m$ time points $(x_1, \ldots, x_m)$ and an associated class label $y$.
    A dataset is a list of $n$ instances, $\textbf{T} = \{(\textbf{X}_1, y_1), \ldots, (\textbf{X}_n, y_n)\}$.
    Given a labelled dataset the objective of a TSC classifier is to predict the class of new input time series using models derived during training.
    We consider the problem of univariate, equal length series in our main results. 
    To demonstrate the robustness of CIF, however, we show results for a range of multivariate datasets and examine an unequal length problem as a case study.

\subsection{Current State of the Art}
\label{curr_sota}
There have been significant recent advances in time series classification. Space restrictions mean we can give only the briefest of summaries of this work.

\textbf{Distance-based} classifiers use time series specific distance functions as the basis of classification. Proximity forest (PF)~\cite{lucas2019proximity} is a classifier that constructs an ensemble of trees built on a randomised selection of time series distance functions. It is currently the most accurate distance based approach, beating the previous best, the Elastic Ensemble~\cite{lines2015time}. 

\textbf{Dictionary-based} classifiers use sliding windows and discretisation to find words in a time series, then classify based on the distribution of these words. WEASEL~\cite{schafer2017fast} and S-BOSS~\cite{large2019time} are dictionary based algorithms that have been shown to be state of the art for this kind of approach~\cite{bagnall2019chapter1}. Both are significantly more accurate than the previous best, the Bags of Symbolic-Fourier-Approximation Symbols (BOSS)~\cite{schafer2015boss}.

\textbf{Shapelet-based algorithms} use phase independent subseries as the basis for classification. Shapelet transform-based algorithms have historically used full enumeration of the shapelets in the train data. This is unnecessary, and often leads to over fitting. The shapelet transform classifier (STC)~\cite{bostrom2017binary} is a shapelet transform based classifier that operates under a maximum search time. In one hour of searching per dataset, it finds a transform that is not significantly worse than full enumeration. STC includes other improvements that make it significantly more accurate than the original transform~\cite{lines2012shapelet}.

\textbf{Spectral} classifiers use features in the frequency domain. The most successful of these to date is RISE~\cite{flynn2019contract}, a tree ensemble that extracts spectral features from intervals.

\textbf{Interval-based} classifiers in the time domain are our focus of interest. The time series forest (TSF)~\cite{deng2013time} is described in detail in Section~\ref{tsf}. Two later interval-based algorithms that were more complex successors to TSF, the Time Series Bag of Features~\cite{baydogan2013bag} and the LSF~\cite{baydogan2016time} have been shown to be no more accurate than TSF on average, and considerably slower~\cite{bagnall2017great}. 

The most successful approaches combine these types of features. HIVE-COTE~\cite{lines2018time} is a meta ensemble of five classifiers, one for each representation: the Elastic Ensemble (EE)~\cite{lines2015time}, BOSS~\cite{schafer2015boss}, STC, RISE and TSF. TS-CHIEF~\cite{shifaz2020ts} is a tree ensemble that uses a combination of distance, dictionary and spectral features. Unsurprisingly, deep learning approaches to time series classification have become very popular. However, many deep learning algorithms do not appear to perform that well~\cite{fawaz2019deep}, at least on the UCR archive which contains a number of datasets that may be too small for effective deep learning training. The counter-example to this and the best current deep learning classifier is InceptionTime~\cite{fawaz2019inceptiontime}.

\subsection{Time Series Forest (TSF)}
\label{tsf}
    TSF~\cite{deng2013time} aims to capture basic summary features from intervals of a time series. For any given time series of length $m$ there are $m(m-1)/2$ possible intervals that can be extracted.
    TSF takes a random forest-like approach to sampling these intervals.
    For each tree, $k$ intervals are randomly selected, each with a random start position and length.  Each interval is summarised by the mean, standard deviation and slope, and the summaries of each interval are concatenated into a single feature vector of length $3k$ for each time series. A decision tree is built on this concatenated feature vector.
    New cases are classified using a majority vote of all trees in the forest. 
    
    The version of TSF used in the bake off~\cite{bagnall2017great} employed the random tree used by random forest. However, the decision tree described in~\cite{deng2013time} has some minor differences to the random tree. It makes no difference in terms of accuracy, but the tree from~\cite{deng2013time}, the time series tree, has advantages in terms of interpretability.    
    The computational complexity of TSF is $O(nlog(n) \cdot m \cdot r)$~\cite{deng2013time}, where $r$ is the number of trees in the forest. 
    
    

    
\subsection{Canonical Time Series Characteristics (catch22)}

catch22~\cite{lubba2019Catch22} is a set of 22 descriptive features for use in time series analysis.
    The motivation for catch22 is to form a concise and informative subset of the time series features from the 7658 contained in the \textit{hctsa} toolbox~\cite{fulcher2017hctsa}. These features could be used in any mining context. However, the process of selection of features and the experimentation presented in~\cite{lubba2019Catch22} is primarily based on classification with the UCR archive. 
    
    From the original 7658 hctsa features, 766 features sensitive to mean and variance were removed due to the fact that the majority of the data in the UCR archive have been normalised. This was further pruned down to 4791 candidates by removing features which cannot be calculated on over 80\% of datasets. This failure is caused by characteristics of the data such as repeating values and negative values. A three step process was used to further reduce the number of features. 
    For each feature, a stratified cross-validation was performed over each dataset in the UCR archive using a decision tree classifier.    Features which performed significantly better than random chance (according to the class-balanced accuracy metric) were retained.
    These significant features were then sorted by their balanced accuracy, and those below a threshold were removed. A hierarchical clustering was performed on the correlation matrix of the remaining features to remove redundancy.
    These clusters were sorted by balanced accuracy and a single feature was selected from the 22 clusters formed, taking into account balanced accuracy results, computational efficiency and interpretability.
    
    The catch22 features cover a wide range of concepts such as basic statistics of time series values, linear correlations, and entropy. 
    The computational complexity for computing the catch22 features is $O(n \cdot m^{1.16})$~\cite{lubba2019Catch22}, with the exponent on the series length found through computational experiments~\cite{lubba2019Catch22}.
For classification, the obvious way to use catch22 is as a transform prior to building a classifier. The reported results~\cite{lubba2019Catch22} are found using an decision tree classifier, although the default implementation in code uses a random forest.

\section{Canonical Interval Forest Classifier (CIF)}
\label{sec:cif}


catch22 summarises time series features very concisely, but classifier's built on catch22 transformation over the whole series are significantly worse than alternative feature based approaches. TSF is very fast and provides diversity that improves HIVE-COTE, but as a stand alone classifier it is significantly worse than two HIVE-COTE components, BOSS and STC. Our research question was whether by replacing the three simple summary features used in TSF with the more descriptive catch22 feature sets we could find a significantly better interval classifier which in turn could lead to an improvement in HIVE-COTE. Our initial approach on the way to forming CIF is to simply replace the feature extraction operations with no further alterations. 
We call this feature-swapped version `hybrid'. The feature set changes from $f(\cdot) = \{mean, stdev, slope\}$ to $f(\cdot) = \{c22f_1, c22f_2, \ldots, c22f_{22}\}$, where $c22f_i$ indicates the $i^{th}$ canonical feature. 


     Experimentation presented in Section~\ref{sec:results} shows hybrid is significantly more accurate than TSF, but this comes at a time overhead. catch22 features are near-linear to compute, but clearly still more expensive overall than calculating just the mean, standard deviation and slope. There are also characteristics of the catch22 features that cause a large time overhead for some data.
     Two features in particular, the positive and negative \textit{`DN\_OutlierInclude'} features, took a very long time to compute on certain intervals of unnormalised datasets, which exhibit large absolute value extremes. Because of this, CIF uses normalised intervals for these two catch22 features. The rest of the features are calculated on the unnormalised intervals, as the absolute values within particular intervals can  hold important discriminatory information for those features that do leverage it acceptably.
    
    
    
    We include the three TSF features along with the catch22 features in CIF, since evidently for some problems these are sufficient by themselves and very cheap to process. We leverage the larger total feature space and inject additional diversity into the ensemble by randomly sampling the 25 features for each tree. This has the added benefit of improving time efficiency. By default we set the number of features subsampled for each a tree, $a$, to 8. We found this value to be the smallest well-performing value on average during our exploratory experiments. Finally, we employ the time series tree~\cite{deng2013time} originally used by TSF rather than the random tree used in the open source Java implementation~\cite{bagnall2017great}. 
    
    TSF was originally designed for univariate time series. We can expand the use of CIF to multivariate problems by expanding the random interval search space, defining an interval as coming from a random dimension, in addition to having random positions and lengths. Consequently, we augment the number of intervals selected per tree to $k=\sqrt{d}\cdot\sqrt{m}$. 
    
    The full training procedure for CIF is described in Algorithm~\ref{cifAlgo}.
    
     \begin{algorithm}[]
        \caption{buildCIF(A list of $n$ cases of length $m$ with $d$ dimensions, ${\bf T}=({\bf X,y})$)}
        \label{cifAlgo}
        \begin{algorithmic}[1]
            \REQUIRE the number of trees, $r$, the number of intervals per tree, $k$, and the number of attributes subsampled per tree, $a$ (default $r=500$, $k=\sqrt{d}\cdot\sqrt{m}$, and $a=8$)
            \STATE Let $\bf{F} = (F_1 \ldots F_r)$ be the trees in the forest
        	\FOR {$i \leftarrow 1$ to $r$}
                \STATE Let $\bf{S}$ be a list of $n$ cases $(s_1 \ldots s_n)$ with $a \cdot k$ attributes
                \STATE Let $\bf{U}$ be a list of $a$ randomly selected attribute indices $(u_1 \ldots u_a)$
        		\FOR {$j \leftarrow 1$ to $k$}
        		    \STATE $b = rand(1,m-3)$
        		    \STATE $l = rand(b+3,m)$
        		    \STATE $o = rand(1,d)$
        		    \FOR {$t \leftarrow 1$ to $n$}
        		        \FOR {$c \leftarrow 1$ to $a$}
        		            \IF{$u_c <= 22$}
        		                \STATE $s_{t,a(j-1)+c} = c22Feature(\bf{u_c},X_{t,o},b,l)$
        		            \ELSE
        		                \STATE $s_{t,a(j-1)+c} = tsfFeature(\bf{u_c},X_{t,o},b,l)$
        		            \ENDIF
        		        \ENDFOR
        		    \ENDFOR
        		\ENDFOR
        		\STATE $F_i.buildTimeSeriesTree([S,y])$
            \ENDFOR
        \end{algorithmic}
    \end{algorithm}

\section{Experimental Methods}
\label{sec:methodology}
    
    When evaluating the predictive performance of algorithms, we make use of the univariate TSC datasets in the UCR archive~\cite{dau2019ucr}.
    Experiments are run on 112 of the 128 datasets in the archive; we have removed data that have unequal length series or contain missing values, since most algorithm implementations are unable to handle these scenarios. 
    We also remove the dataset Fungi as it only provides a single train case for each class, making parameter optimisations difficult. 
    A summary of those of the 112 datasets that are new to the archive can be found on the accompanying website\footnote{https://sites.google.com/view/icdm-cif/home}.   The UCR archive provides a default split into train and test sets. 
    We resample each dataset 30 times in a random stratified manner. 
    When presenting the performance of a classifier on a dataset, we use the average score across 30 resamples. 
    However, fold 0 is always the original provided split for ease of comparison to other results. 
    The resampling and all classifier parameters are seeded by the dataset fold index, and as such all data, results and analysis are reproducible.
    For our multivariate experiments we use the 26 equal length datasets from the UEA multivariate TSC archive~\cite{bagnall2018uea}, presenting results on the default train and test split only.

 We use two open source software tools that contain implementations provided by the original algorithm designers. \texttt{tsml}\footnote{https://github.com/uea-machine-learning/tsml} is a Weka compatible time series machine learning toolkit that contains implementations of the majority of the existing algorithms we have evaluated. \texttt{sktime}\footnote{https://github.com/alan-turing-institute/sktime} is scikit-learn compatible toolkit for time series with a deep learning variant called \texttt{sktime-dl}\footnote{https://github.com/sktime/sktime-dl}. Our adaptations of these toolkits and code to reproduce all experimental results are available on the accompanying website.

    When comparing two classifiers over multiple data sets, we use a pairwise Wilcoxon signed-rank test on the scores averaged over resamples.  We present results of multiple classifiers over multiple data sets using an adaptation of critical difference diagrams~\cite{demvsar2006statistical}, with the change that all classifiers are compared with pairwise Wilcoxon signed-rank tests, and cliques formed using the Holm correction~\cite{garcia2008extension} rather than the post-hoc Nemenyi test originally used by~\cite{demvsar2006statistical}. We assess classifier performance with accuracy. We were constrained to seven days computation for a single experiment (i.e. a single resample of a data set with a particular classifier).

\section{Results} 
\label{sec:results}
Our experiments are designed to answer the following questions. 

\noindent \textbf{What is state of the art in TSC?} In Section~\ref{sec:sota} we present results for eight recently proposed classifiers on the expanded UCR archive. We show why the interval-based representation needs improving.

\noindent \textbf{Does CIF improve over TSF, and how does it compare to the best classifiers of alternative single representation?} We demonstrate the improvements made and compare CIF to the best in class for the other representations in Section~\ref{sec:cif_acc}. We test the aptitude of CIF for multivarate data in Section~\ref{sec:multivariate}.

\noindent \textbf{Does CIF improve HIVE-COTE?} In  Section~\ref{sec:hive-cote} we compare HIVE-COTE with CIF to state of the art. 

\noindent \textbf{How efficient is CIF in relation to other algorithms?} 
In Section~\ref{sec:tsefficiency} we quantify the time and space efficiency of CIF and propose a more efficient mechanism for estimating test accuracy for CIF's use in HIVE-COTE. 

\subsection{Recent TSC Algorithms}
\label{sec:sota}

\begin{figure}
        	\centering
            \includegraphics[width=0.98\linewidth,trim={0 5cm 0 4cm},clip]{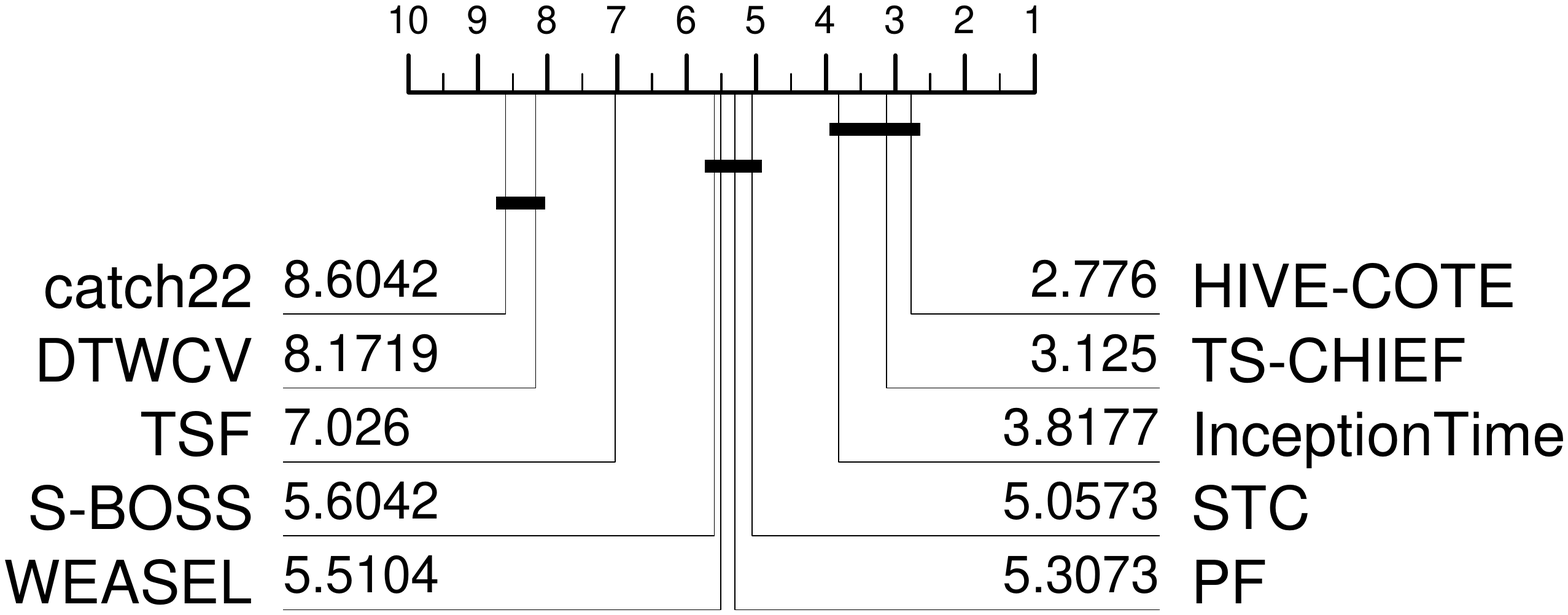}
            \caption{Critical difference diagram for rankings by classification accuracy of eight recently proposed TSC algorithms, TSF and DTWCV.}
            \label{fig:miniBakeoffAcc}
        \end{figure}

In Section~\ref{sec:background} we described eight recently proposed algorithms for TSC. We present the critical difference diagram for the 97 problems all algorithms could complete for accuracy (Figure~\ref{fig:miniBakeoffAcc}).
Due to space limitations, we only additionally report area under the receiver operating characteristic (AUROC) for our main contributions. The full results for classifiers shown in our results for both these metrics, as well as balanced accuracy and F1 score, are provided on the accompanying website. 
Most algorithms finished at least 109 of the 112. HIVE-COTE only completed 97. This was caused by the Elastic Ensemble (EE) component. EE performs a ten fold cross validation to find the HIVE-COTE weight, which slows it down considerably.    
We include DTWCV for reference and TSF as it is relevant to later results. The number adjacent to each algorithm indicates the average rank over all problems (lower is better). Solid bars represent cliques within which there is no significant difference. Experiments were performed with the parameters used in the paper of origin of each classifier, with the exception of STC which uses a max search time of 1 hour instead of full enumeration. These settings are described on the accompanying website.

Our first observation is that DTWCV is not competitive with modern TSC algorithms. On average, S-BOSS is 7\% more accurate per problem, and HIVE-COTE is 10.5\% better. There are obviously still scenarios where DTW will be useful for classification and it has much broader applications. However, we think that it is no longer hard to beat and, in isolation, is not a valid benchmark for comparisons based on classification accuracy. There are three clear cliques evident in the accuracy ranks shown in Figure~\ref{fig:miniBakeoffAcc}. 
catch22 is the only algorithm not significantly better than DTWCV. Like DTW, catch22 has applications beyond classification. However, using it as a whole series transformation prior to classification is not very effective. TSF is significantly better than both, but worse than the first clique of modern classifiers: S-BOSS, WEASEL, PF and STC. Each of these algorithms rely on a single representation/transformation.
The top clique, InceptionTime, TS-CHIEF and HIVE-COTE, contains algorithms that use combinations of representations. 
Although the hybrid approaches are the most accurate, it is, in our opinion, worthwhile researching new algorithms based on single representations. Some data will be best approached by one representation based on the domain specific problem definition. The four single representation algorithms evaluated all represent a significant improvement on the previous best of class that are currently used in HIVE-COTE. Our objective is to produce a new interval-based algorithm that represents a significant improvement over TSF. 

\subsection{CIF as a standalone classifier}
\label{sec:cif_acc}

\begin{figure}
        	\centering
            \includegraphics[width=0.98\linewidth]{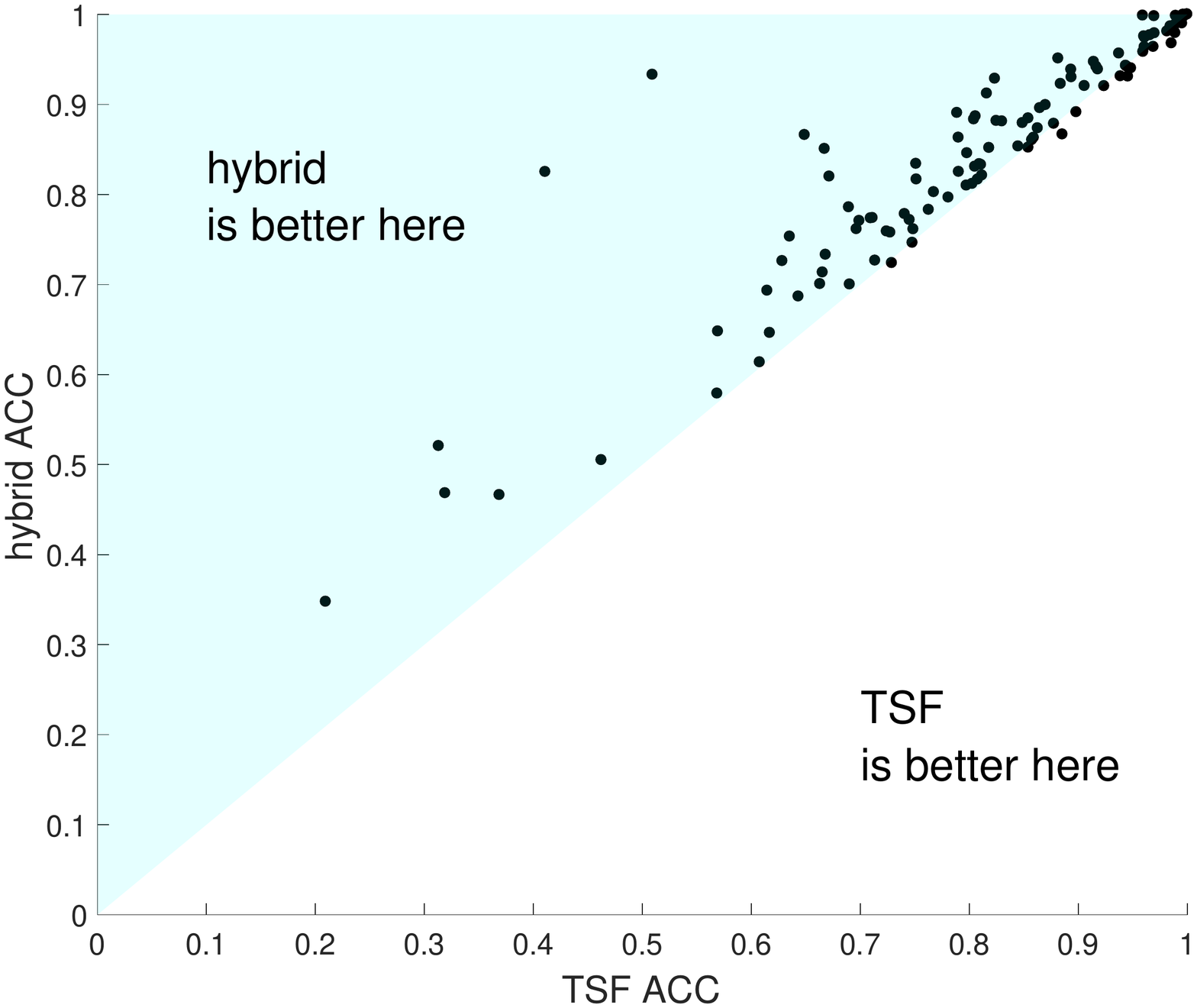}
            \caption{Scatter plot of TSF vs hybrid on 109 problems. Hybrid wins on 87 problems, ties on 3, and loses on 19 problems.}
            \label{fig:tsfVshybrid}
        \end{figure}

\begin{figure}
        	\centering
            \includegraphics[width=0.98\linewidth,trim={0 11cm 0 4cm},clip]{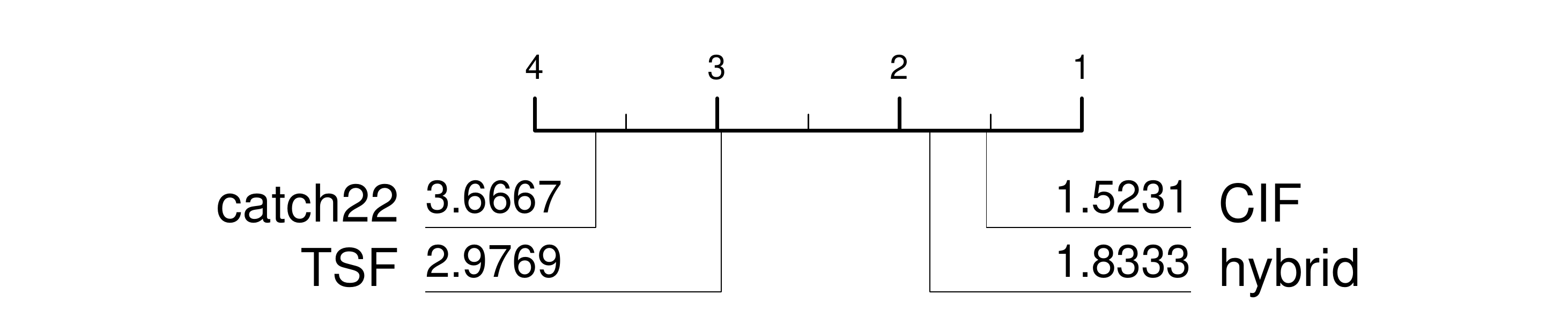}
               \caption{Accuracy rank comparison for three interval classifiers and catch22 on 109 UCR problems.}
               \label{fig:intervalVariants}
        \end{figure}

 Our first experiment compares the hybrid classifier that uses all of the catch22 features instead of the three TSF features. This serves as a basic test of concept.  Figure~\ref{fig:tsfVshybrid} presents a pairwise scatter diagram comparing accuracies of TSF and hybrid.
The hybrid is significantly better than TSF. If we examine the resamples on each dataset and perform a paired sample t-test, we find that the hybrid is significantly better on 76 datasets and significantly worse on just one. There is improvement from using the catch22 features with TSF. However, the simple usage of catch22 presents some problems. The hybrid is drastically slower than TSF on some problems. There are only results 109 problems in Figure~\ref{fig:tsfVshybrid} because hybrid failed to complete three problems within 7 days. CIF is designed to improve efficiency (see Section~\ref{sec:efficiency}) but also inject further diversity. 
Figure~\ref{fig:intervalVariants} shows the rank data for CIF, catch22, TSF and the hybrid. 
It clearly demonstrates that CIF improves the accuracy of the simple catch22-TSF hybrid. The actual accuracy improvements over TSF and catch22 are not small. The average improvement of CIF over all data sets is 4.56\% against TSF and 6.34\% vs catch22. CIF has higher accuracy on 99 of the 109 problems.

\begin{figure}
        	\centering
            \includegraphics[width=0.98\linewidth,trim={0 9cm 0 5cm},clip]{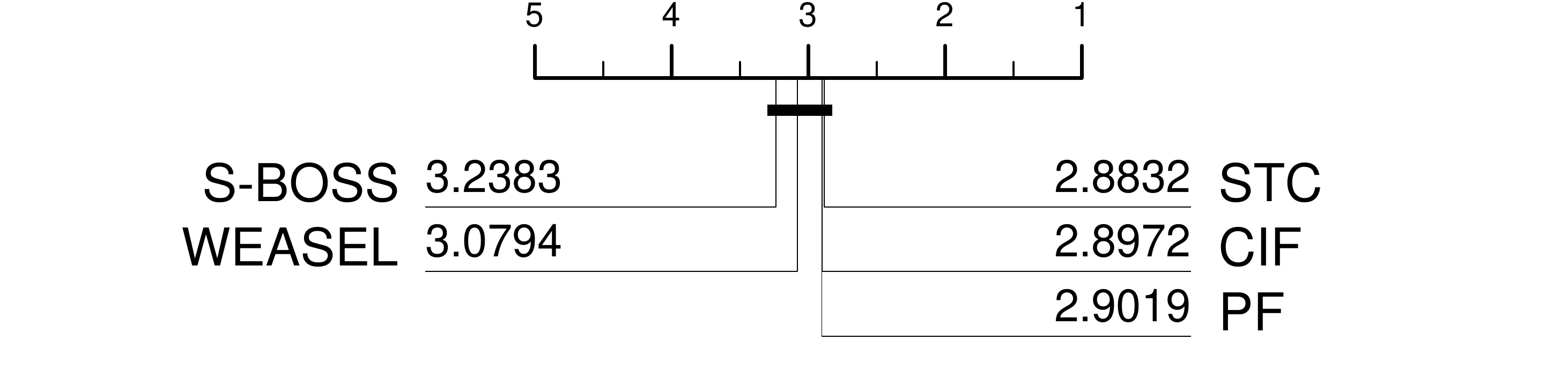}
               \caption{Accuracy rank comparison for CIF and latest single representation classifiers. }
               \label{fig:bestInClassACC}
        \end{figure}
\begin{figure}
        	\centering
            \includegraphics[width=0.98\linewidth,trim={0 9cm 0 5cm},clip]{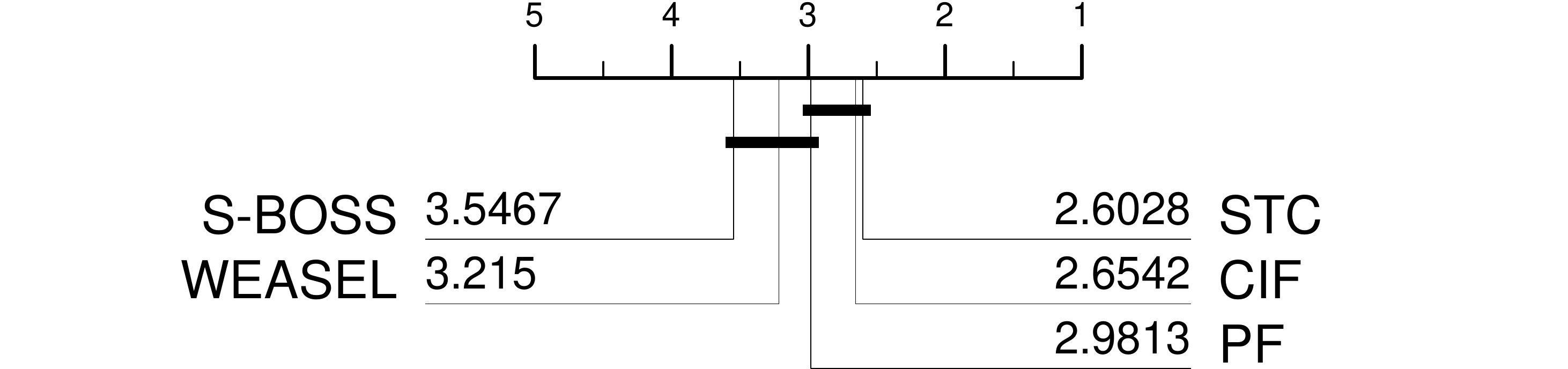}
               \caption{AUROC rank comparison for CIF and latest single representation classifiers. }
               \label{fig:bestInClassAUROC}
        \end{figure}

CIF is the best in class for the interval-based classifiers. Figure~\ref{fig:bestInClassACC} compares CIF to the latest single representation classifiers: PF; STC; WEASEL; and S-BOSS. There is no significant difference between these five classifiers. Ranking based on the AUROC metric (Figure~\ref{fig:bestInClassAUROC}), CIF performs significantly better then S-BOSS and WEASEL. We conclude it achieves our goal of developing an interval-based classifier that is at least as good as other algorithms based on a single representation.
        
        \begin{figure}
        	\centering
            \includegraphics[width=0.98\linewidth]{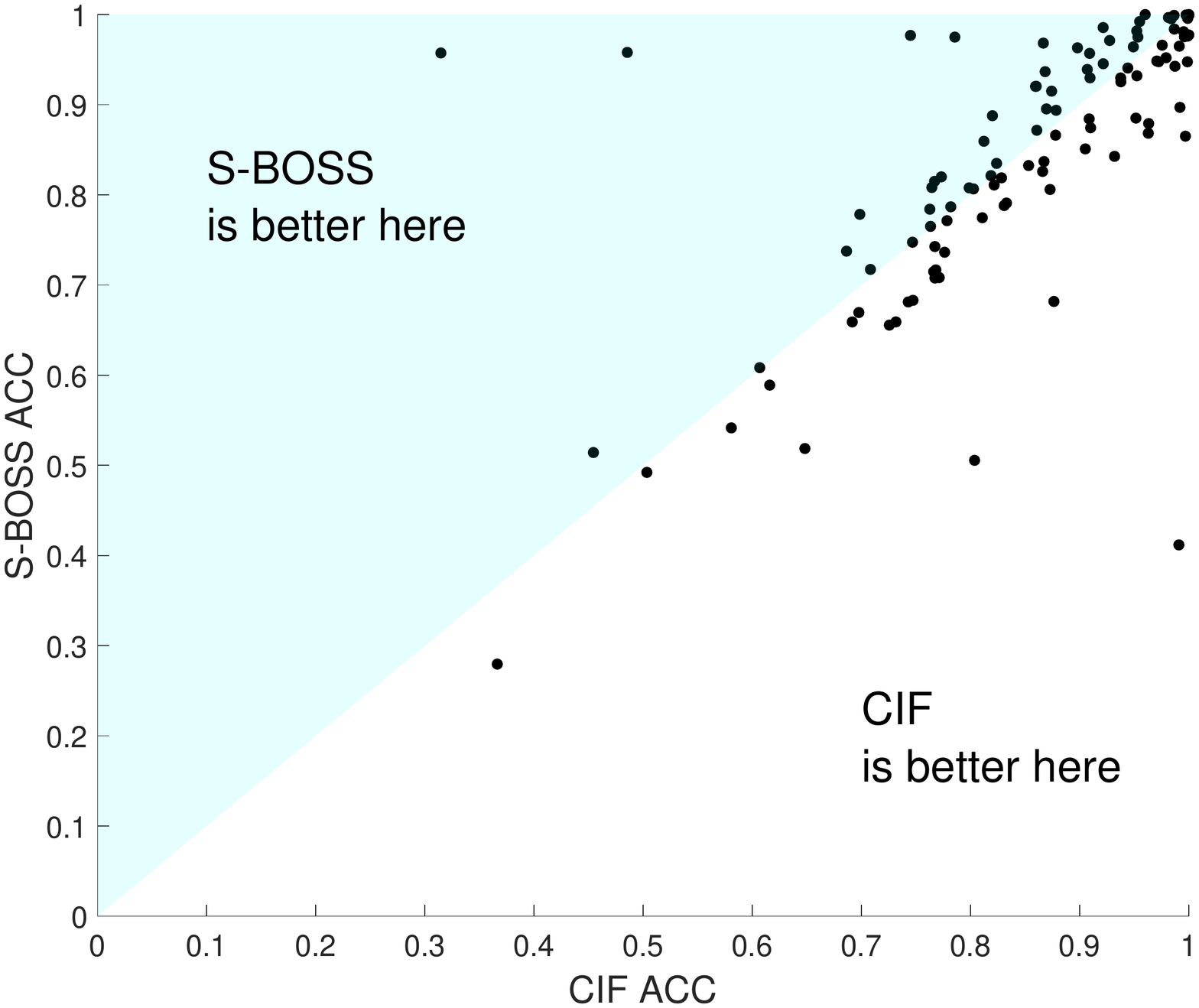}
            \caption{Scatter plot of S-BOSS and CIF. CIF wins on 59 problems, ties on 1, and loses on 47 problems.}
            \label{fig:cifScatter}
        \end{figure}
 
\subsection{CIF on multivariate data}
\label{sec:multivariate}

        We have demonstrated CIF's performance on univariate data. Strong performance here does not necessarily transfer to the multivariate case, however. To demonstrate this we compare to against a range of multivariate classifiers on the UEA multivariate archive.
        DTW is still a reasonable benchmark in the multivariate case, using the different strategies for generalisation described in~\cite{shokoohi2017generalizing}. $DTW_I$ assumes independence between dimensions, $DTW_D$ assumes dependence, while $DTW_A$ adaptively selects between the two on an instance by instance basis.
        The generalised random shapelet forest (gRSF)~\cite{karlsson2016generalized} constructs an ensemble of shapelet trees, randomly selecting the dimension each shapelet is generated from.
        Also included are TSF and cBOSS~\cite{middlehurst2019scalable}, a more scalable version of the HIVE-COTE dictionary component BOSS~\cite{schafer2015boss}. For both classifiers a model is trained for each dimension independently, which are then ensembled with new cases being classified using majority vote. We call these $TSF_I$ and $CBOSS_I$.
        We attempted to obtain results for two other classifiers, however, we have needed to omit them: the deep learning method TapNet~\cite{zhang20tapnet}, for which we could not satisfactorily reproduce the results for using the authors code; and WEASEL+MUSE~\cite{schafer2017multivariate}, which required more than 500GB on 4 problems.
        
\begin{figure}
        	\centering
            \includegraphics[width=0.98\linewidth,trim={0 8cm 0 5cm},clip]{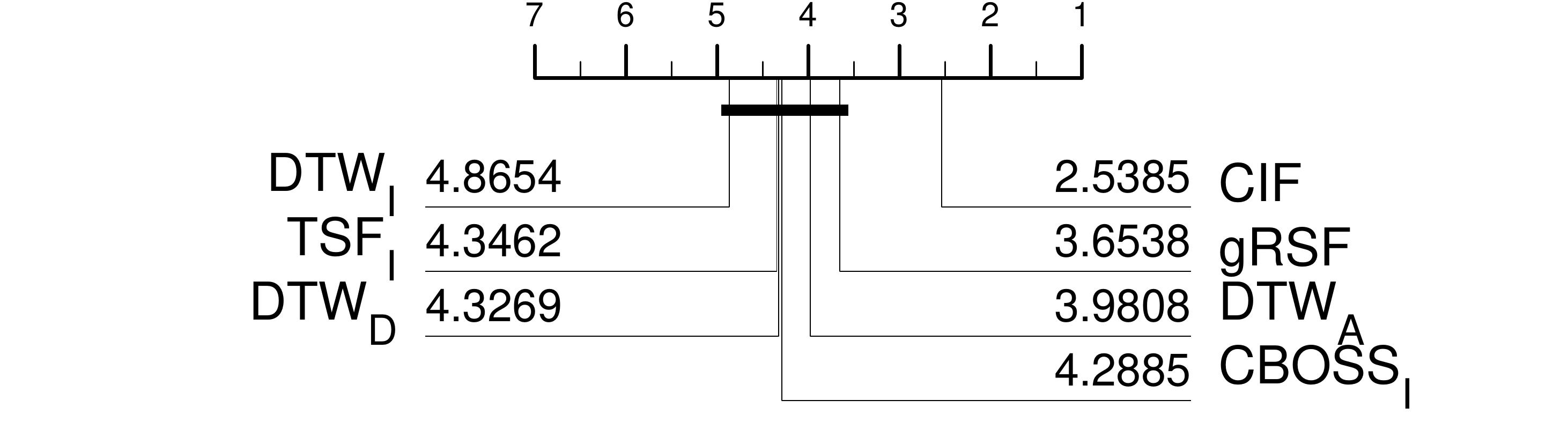}
            \caption{Accuracy rank comparison for CIF and multivariate classifiers on 26 equal length UEA datasets.}
            \label{fig:multivariate}
        \end{figure}

        Figure~\ref{fig:multivariate} shows the critical difference diagram for CIF and the multivariate classifiers presented. DTW remains a good benchmark for multivariate data. TSF and cBOSS are significantly better then univariate DTW, this does not translate to the multivariate case. However, CIF performs significantly better than the other algorithms used in the analysis. Whilst the comparison is not exhaustive, it does  
        demonstrate that it is easy and intuitive to generalise CIF effectively for the multivariate case. 
        

\subsection{CIF as a HIVE-COTE component}
\label{sec:hive-cote}

\begin{figure}
        	\centering
            \includegraphics[width=0.98\linewidth,trim={0 10cm 0 5cm},clip]{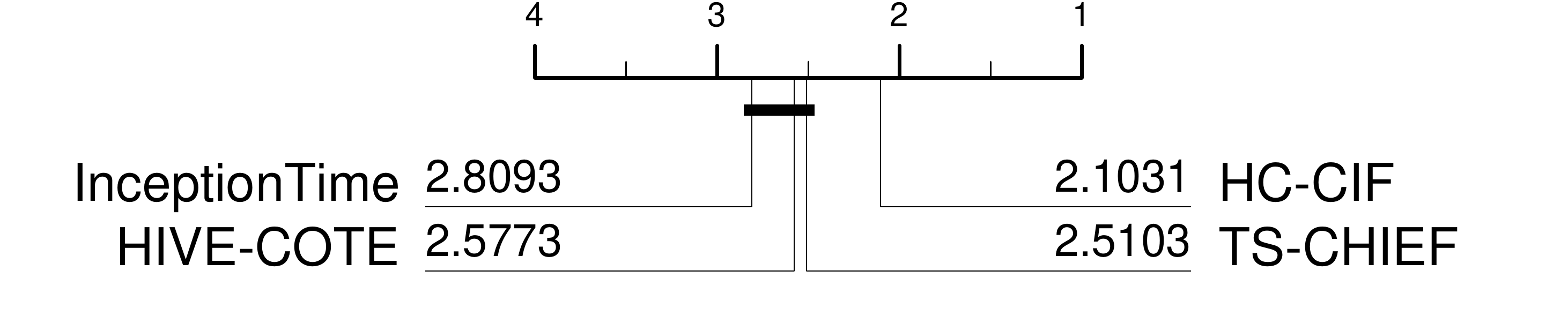}
               \caption{Accuracy rank of three state of the art classifiers and HIVE-COTE with CIF replacing TSF (HC-CIF) on 30 resamples of 97 UCR classification problems.}
               \label{fig:sotaACC}
        \end{figure}

\begin{figure}
        	\centering
            \includegraphics[width=0.98\linewidth,trim={0 11cm 0 5cm},clip]{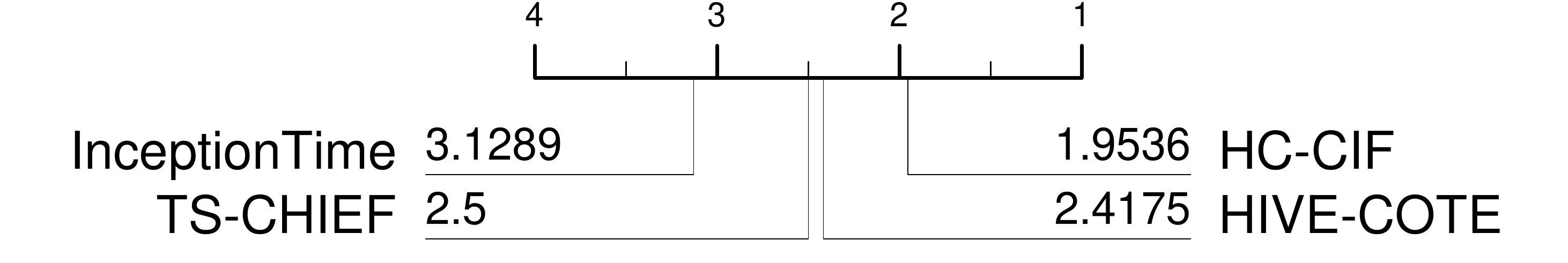}
               \caption{AUROC rank of three state of the art classifiers and HIVE-COTE with CIF replacing TSF (HC-CIF) on 30 resamples of 97 UCR classification problems.}
               \label{fig:sotaAUROC}
        \end{figure}

CIF is not better than other single representation classifiers, so why do we need CIF if we already have STC, PF, S-BOSS and WEASEL? It is not just a matter of the efficiency benefits (discussed in Section~\ref{sec:efficiency}). Despite there being no overall difference between these single representation classifiers, there is still significant diversity over specific problems.  Figure~\ref{fig:cifScatter} shows the accuracy scatter plot for CIF and S-BOSS. Each representation specialises on different data characteristics, causing large differences in accuracy. The same pattern can be observed with the other classifiers, for example, CIF and STC vary significantly over data. STC beats CIF by over 40\% on PigCVP and PigAirwayPressure, whereas the opposite is true with TwoPatterns. There are 12 problems with a difference in accuracy of 10\% or more. This diversity is what helps HIVE-COTE perform so well. If CIF can improve HIVE-COTE, then it is further evidence as to the value of improving interval based approaches. To test this we replace TSF with CIF and rerun HIVE-COTE, whilst leaving the other components the same. To differentiate between the old and new variants of HIVE-COTE, we shall call them HIVE-COTE and HC-CIF respectively. Figure~\ref{fig:sotaACC} gives the accuracy ranks of the state of the art classifiers and HC-CIF. HC-CIF is significantly more accurate than all three. This is also the case for AUROC (Figure~\ref{fig:sotaAUROC}). 
We conclude that HC-CIF is the most accurate classifier to date (to our knowledge) on the UCR time series classification archive, and as such represents a genuine advance in state of the art for this field. We stress we have done nothing to HIVE-COTE except replace TSF with CIF.    

\begin{figure}
        	\centering
            \includegraphics[width=0.98\linewidth]{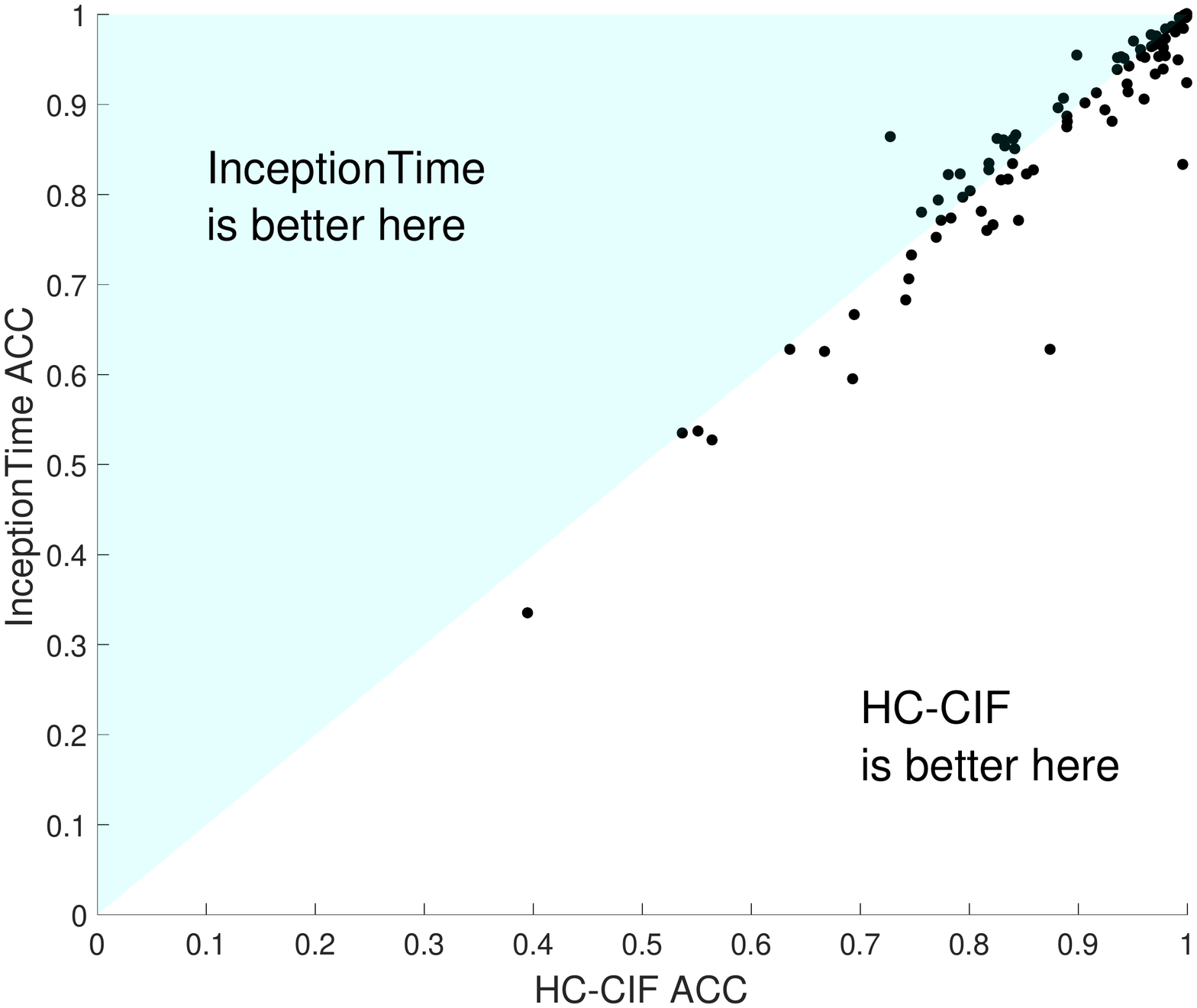}
            \caption{Scatter plot of InceptionTime and HC-CIF. HC-CIF wins on 68 problems, ties on 5, and loses on 36 problems.}
            \label{fig:inception}
        \end{figure}

Figure~\ref{fig:inception} show the scatter plots for HC-CIF against InceptionTime, and includes the win/tie/lose statistics for information. There is still wide variation, but the improvement from HC-CIF is visually apparent. HC-CIF has, on average, 1.56\% higher accuracy than InceptionTime.  

\section{CIF EFFICIENCY AND USABILITY}
\label{sec:efficiency}

\subsection{Time and space efficiency}
\label{sec:tsefficiency}

\begin{figure}
        	\centering
            \includegraphics[width=0.98\linewidth]{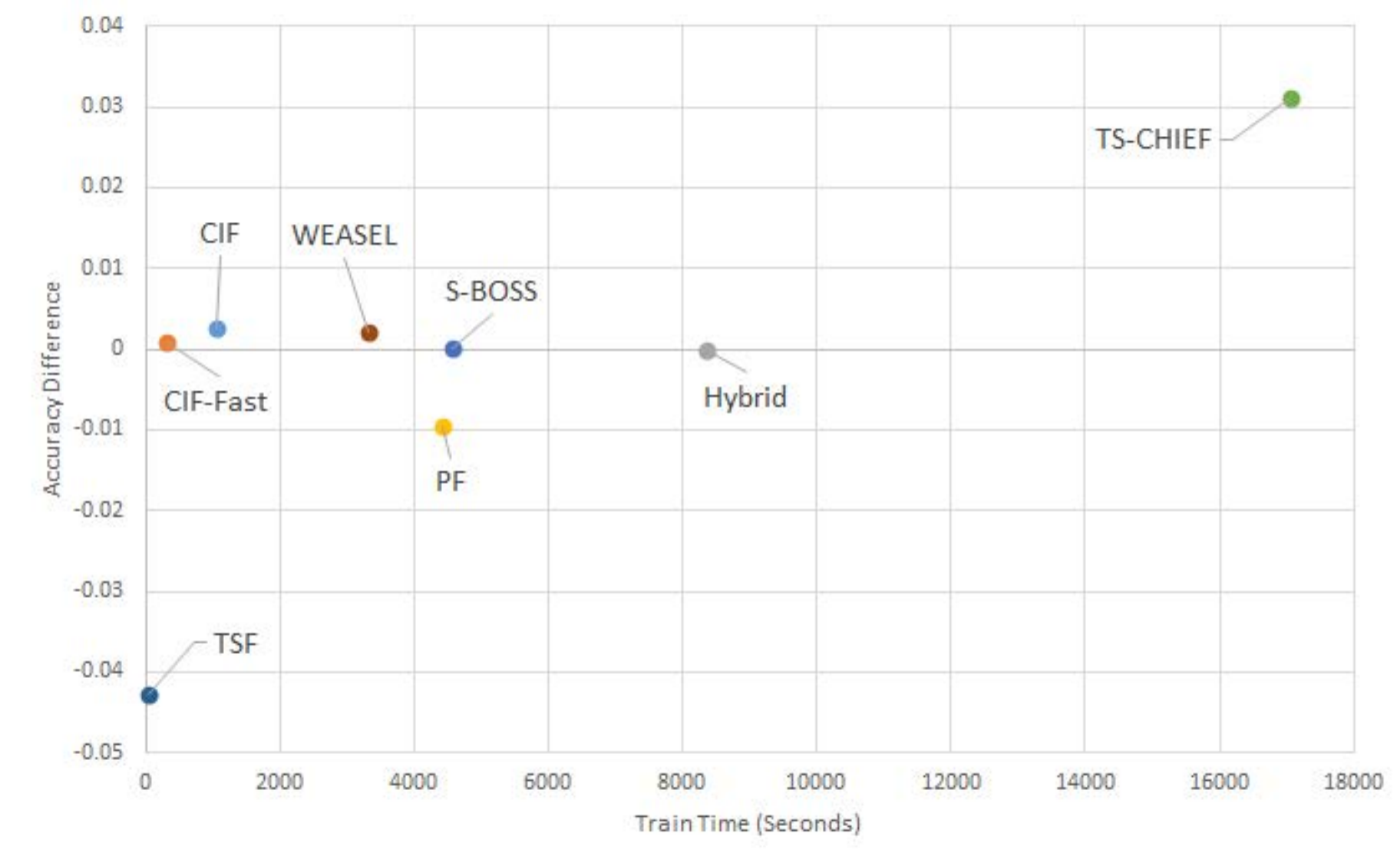}
            \caption{A comparison of classifiers in terms of accuracy and train time. The y-axis shows the average difference in accuracy from the default hybrid approach. Time and accuracy are averaged over 104 UCR problems. }
            \label{fig:cifTimeAcc}
        \end{figure}

CIF is significantly more accurate than TSF, but at what cost in terms of time and memory? Figure~\ref{fig:cifTimeAcc} plots the average difference in accuracy from a baseline (hybrid) against the average run time on the UCR data for CIF and 6 algorithms used in our previous comparisons. All experiments were conducted in a single thread, even though some such as PF and TS-CHIEF can employ multiple threads. This is to make sure comparisons are fair. 
        
At the extremes, we see TSF is by far the fastest, but the least accurate; and that TS-CHIEF is the most accurate, but the slowest. CIF is approximately eight times faster than the hybrid, and is the fastest of the single representation classifiers displayed in Figure~\ref{fig:bestInClassACC}, between which there is no significant difference in accuracy. CIF-Fast is a reduced form of CIF with fewer trees (250) and intervals per tree ($\sqrt{m}^{0.85}$). We include it to demonstrate that, if the problem is very large, CIF can be configured to be faster with little loss of accuracy. To further improve usability, we provide a contracted version of CIF in line with other HIVE-COTE components~\cite{bostrom2017binary,flynn2019contract,middlehurst2019scalable} that adds trees to the ensemble incrementally. The ability to set a fixed time to train until allows a degree of certainty for time sensitive training requirements, such as clusters with a maximum job time limit.

\begin{figure}
        	\centering
            \includegraphics[width=0.98\linewidth,trim={5cm 2.5cm 3cm 1cm},clip]{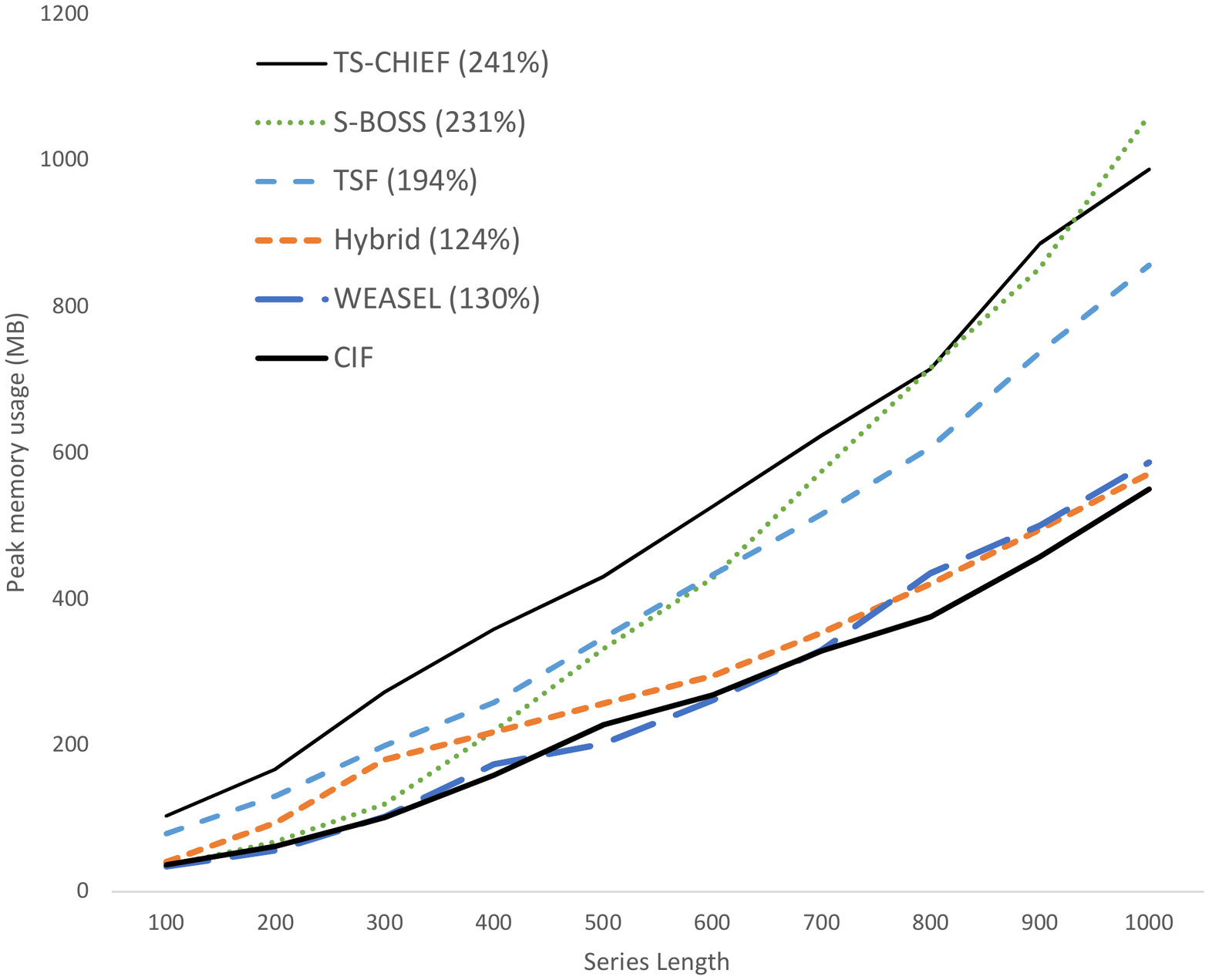}
               \caption{Memory usage of six classifiers for varying series length. Figures in brackets are the overall percentage of memory used in comparison to CIF.}
               \label{fig:mem}
        \end{figure}

Figure~\ref{fig:mem} characterises the memory footprint of six algorithms using simulated data. We fix the number of cases and increase series length. TS-CHIEF and S-BOSS require the most memory, (241\% and 231\% of that required by CIF respectively). Both employ dictionary-based nearest neighbour classifiers, which require storing representations of the whole dataset. TSF memory usage was a surprise. We can explain the relatively high memory usage by reference to the implementation. The Weka random tree it employs makes a local copy of the data for every tree. This data is not stored after the build phase, but it seems the Java garbage collection is slow in deallocating the memory. The hybrid uses slightly more memory than CIF due to the increased size of its feature space.   

        \begin{figure}
        	\centering
            \includegraphics[width=0.98\linewidth,trim={0cm 0.5cm 0cm 0.5cm},clip]{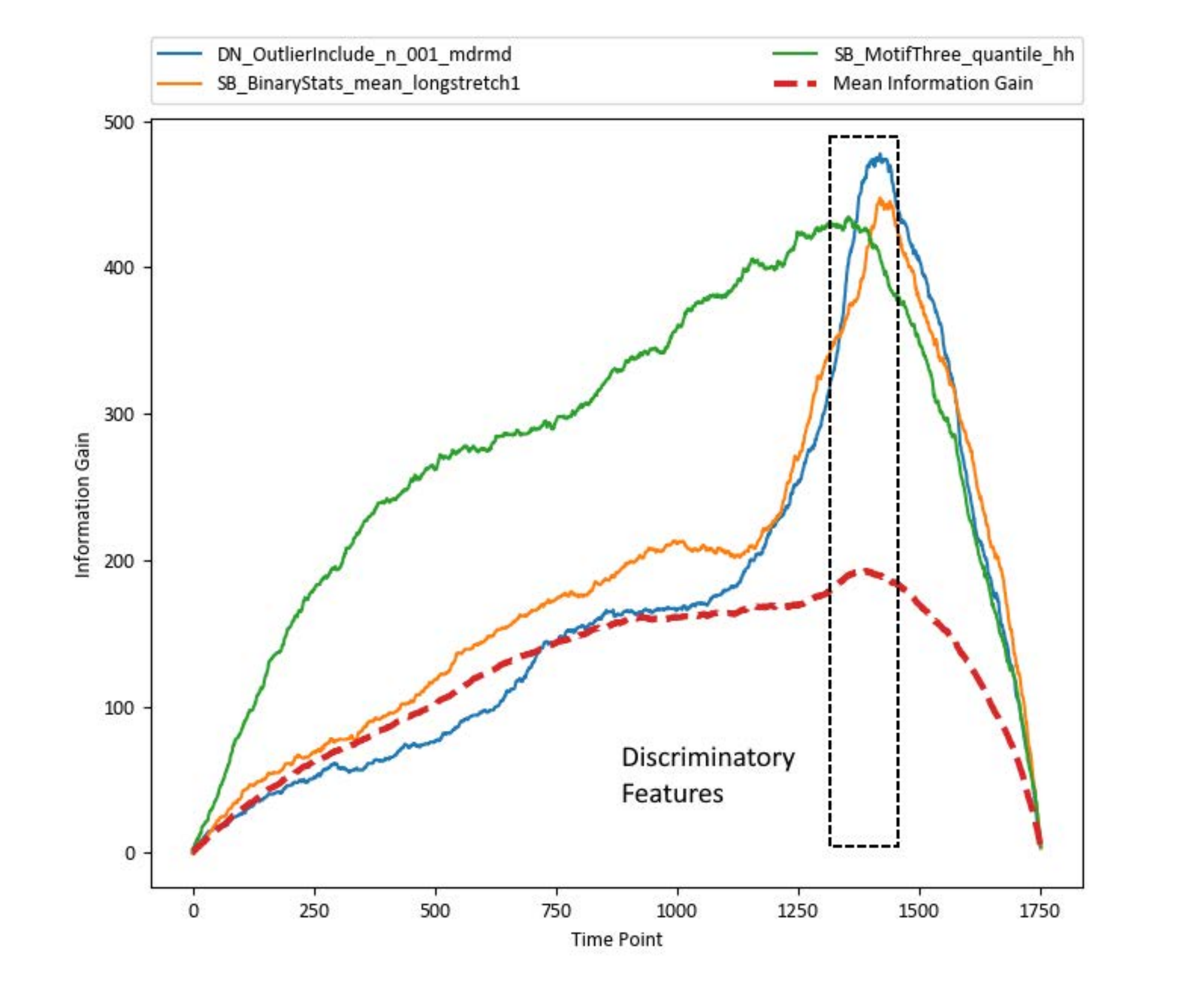}
            \caption{Temporal importance curves for the EthanolLevel problem. Higher values indicate greater importance for classification. The dotted box indicate the most important region as specified by a domain expert (see Figure~\ref{fig:ethanolExample}).}
            \label{fig:ethanolTIC}
        \end{figure}
        
    \subsection{Model Visualisation}
\label{sec:vis}
        
    One benefit of TSF is that it can be used to evaluate the relevance of regions of the time series. Temporal importance curves~\cite{deng2013time} are a visualisation method that displays the importance of time series intervals to classification.
    To achieve this, the information gain of each node's splitting attribute is collected for all trees in the forest.
    Each attribute the trees are built on corresponds to a summary feature and a time interval. 
    A curve is created for each summary feature, adding the gain from any node split to its curve at the time points covered by its interval.
    We can adapt this algorithm for CIF, although displaying all 25 features can be too cluttered. We display the top $v=3$ features by max gain over all time points, as these have effectively been discovered to be the most important, in addition to the mean of all features for each time point.
    
    Figure~\ref{fig:ethanolTIC} displays the CIF temporal importance curves for the EthanolLevel problem shown previously in Figure~\ref{fig:ethanolExample}. It demonstrates that the most important region for classification is the same region identified by a domain expert.  
    While by default CIF subsamples features per tree, we have not done this for the above as it can have a large impact on the production of these curves. 
    The temporal importance curves can be generated for model visualisation and are still meaningful while subsampling. However, randomly restricting access to features, and therefore promoting weaker features that would otherwise have lost to more consistently informative ones when splitting, may not give the best representation to perform a higher level analysis of a time series problem. The goals of maximising accuracy, minimising computational cost, and maximising interpretability, often run counter to each other. 
    
\subsection{Out-of-bag error estimates}
\label{sec:oob}

CIF significantly improves HIVE-COTE. However, HIVE-COTE requires an estimate of the test accuracy from the train data for its weighting scheme.  For the results presented in Section~\ref{sec:cif}, we found these estimates through ten fold cross validation. This cross validation is not included in the timings shown in Figure~\ref{fig:cifTimeAcc}. This extra time overhead is one of the major disadvantages of HIVE-COTE, since it increases the run time of the base classifiers by an order of magnitude. Hence, we investigate other ways of estimating the test accuracy.

 One alternative is to mimic random forest and use bagging with CIF. This means the out-of-bag accuracy can be used for the weight for HIVE-COTE. Unfortunately, our experience with bagging for other tree based ensembles for time series data has shown it makes the final classifier significantly worse than building on the whole train data. 
 Given we wish to build the final model on the full data for testing, the question is whether there is any difference using bagging estimates or cross validation estimates of the full model test accuracy, and whether there is any resulting impact on HIVE-COTE. Table~\ref{tab:bag} shows the difference in actual test accuracy and that predicted from the train data. Cross validation under estimates test accuracy by 2.86\% on average. Bagging has lower error when predicting the test accuracy of a bagged classifier, but the actual test accuracy is significantly lower. Our compromise of using bagging to estimate test accuracy but the full model to predict new cases allows us to achieve lower estimation error than CV approximately five times faster.   
We have verified that using the bagged estimates within HIVE-COTE to weight the CIF module's output does not make any overall significant difference to HIVE-COTE performance.    

        \begin{table}[h]
            \centering
			\caption{Performance in estimating test error by various means for CIF, averaged over 112 UCR datasets.}
			\begin{tabular}{l c c c} 
			\cline{2-4}
&	CV+Full	& Bagging	& Bagging+Full \\ \hline
Actual Test Acc &	84.61\%	& 83.46\%	& 84.61\% \\
Estimated Test Acc	& 81.75\%	& 82.10\%	& 82.10\%\\
Difference & 2.86\%	& 1.37\% & 2.52\%\\ \hline
\end{tabular}
			\label{tab:bag}
\end{table}
    
\section{Case Study}
\label{sec:usecase}

    We demonstrate the usefulness of CIF by using the three Asphalt datasets first presented in~\cite{souza2018asphalt}. The problem involves predicting the condition of a road based on motion data. Data is recorded on a smart phone installed inside a vehicle using a flexible suction holder. This offers the potential for automated monitoring and assessment of road conditions leading to earlier and less costly interventions with faults. 
    The Android application \textit{Asfault}~\cite{souza2017towards} was used to collect accelerometer data in the form of the three physical axes, latitude, longitude, and velocity from GPS ($A_x, A_y, A_z$). These axes are converted into a univariate time series that represents the acceleration magnitude ($A_m$). This magnitude forms the time series used in each problem.    A sampling rate of 100 Hz was used for each time series.
    These datasets cover three separate problems related to road surface conditions.
    Asphalt-Regularity looks at road deterioration using driver comfort as a metric. A road is classified as regular or deteriorated. 
    The Asphalt-PavementType problem is to classify the surface type of the road as either dirt, cobblestone or asphalt. The Asphalt-Obstacles problem is to classify whether the vehicle is crossing one of a set of common road obstacles: speed bumps; vertical patches; raised pavement markers; and raised crosswalks.
    Table~\ref{tab:asphaltStats} presents basic summary stats for the datasets. The series are unequal length, centred around zero, but not normalised. 
    
        \begin{table}[h]
            \centering
			\caption{Summary information for the Asphalt data presented in~\cite{souza2018asphalt}.}
			\begin{tabular}{m{0.29\linewidth} m{0.17\linewidth} m{0.22\linewidth} m{0.15\linewidth}}
				\cline{2-4}
				& Asphalt-Regularity & Asphalt-PavementType & Asphalt-Obstacles \\ \hline
                Train Size & 751 & 1055 & 390 \\
                Test Size & 751 & 1056 & 391 \\
                Min Series Length & 66 & 66 & 111 \\
                Max Series Length & 4201 & 2371 & 736 \\
                No. Classes & 2 & 3 & 4 \\ \hline
			\end{tabular}
			\label{tab:asphaltStats}
		\end{table}
    
    1-NN classifiers were used for this problem in~\cite{souza2018asphalt}, each using a range of elastic distance measures combined with complexity invariant distance (CID)~\cite{batista2014cid}. We compare against the best reported results for both the univariate and multivariate versions. 
    Many of the classifiers we test are unable to handle series of different length. A common initial strategy for dealing with unequal length series is to zero pad the data so that all series are the same length as the longest. 
    Whilst standard, this approach can introduce features that confound classifiers, particularly if the padding is extreme, as in this case.   
    We believe interval-based classifiers such as CIF have an inherent tolerance to issues caused by padding. Experiments are conducted using the same methods described in Section~\ref{sec:methodology}.
		
		\begin{table}[h]
            \centering
			\caption{Classifier accuracy on three asphalt data sets over 30 resamples. Best results from~\cite{souza2018asphalt} were averaged over 5 2-fold cross-validations}
			\begin{tabular}{m{0.29\linewidth} m{0.17\linewidth} m{0.22\linewidth} m{0.15\linewidth}}
			\cline{2-4}
				& Asphalt-Regularity & Asphalt-PavementType & Asphalt-Obstacles \\ \hline
                Best~\cite{souza2018asphalt} $A_m$ & 0.9648 & 0.8066 & 0.8113 \\
                Best~\cite{souza2018asphalt} $A_xA_yA_z$ & 0.9848 & 0.8827 & 0.7944 \\
                CIF $A_m$     & \textbf{0.9863}   & 0.9015 & 0.8257 \\
                CIF $A_xA_yA_z$    & 0.9613  & 0.8594 & 0.7877 \\
                TSF     & 0.9736                & 0.8647 & 0.7982 \\
                S-BOSS  & 0.8154 & 0.6223 & 0.8335 \\
                STC     & 0.9251 & 0.8173 & 0.8209 \\
                RISE & 0.5077 & 0.3864 & 0.4038 \\
                PF      & 0.9836 & 0.8775 & 0.8891 \\
                TS-CHIEF & N/A & N/A & 0.8858 \\
                HC-CIF  & 0.9482 & 0.8704 & 0.8408 \\
                InceptionTime & 0.9860 & \textbf{0.9369} & \textbf{0.9100} \\ \hline
			\end{tabular}
			\label{tab:asphaltResults}
		\end{table}
		
	Results for the three asphalt datasets are shown in table~\ref{tab:asphaltResults}. 
	The  best performing single representation classifier is CIF and the best overall is InceptionTime. CIF beats HC-CIF (HIVE-COTE with CIF) on two of the three problems. Analysis of HC-CIF shows that the performance of the RISE component reduced the overall accuracy of the meta ensemble.  RISE  takes spectral features on a single interval per tree, which supports our belief that the multiple intervals taken by CIF mitigate the impact of padding. 	Furthermore, we omit EE from this version of HIVE-COTE because it could not complete within our time constraints. TS-CHIEF did not complete two of the problems within our 7 day limit. 
        \begin{figure}
        	\centering
            \includegraphics[width=0.98\linewidth,trim={0cm 1.49cm 0cm 1.5cm},clip]{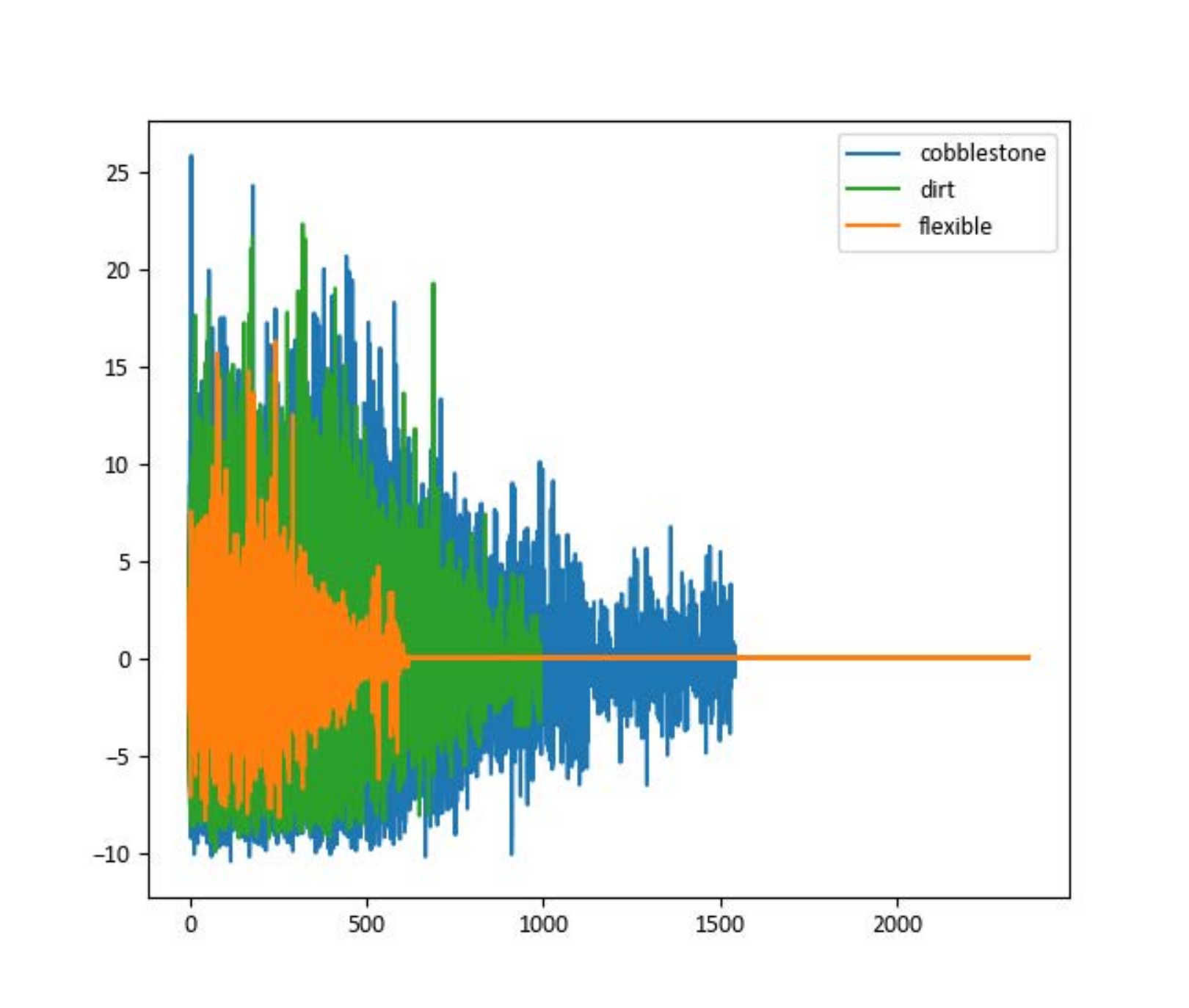}
            \includegraphics[width=0.98\linewidth,trim={0cm 0.5cm 0cm 0.5cm},clip]{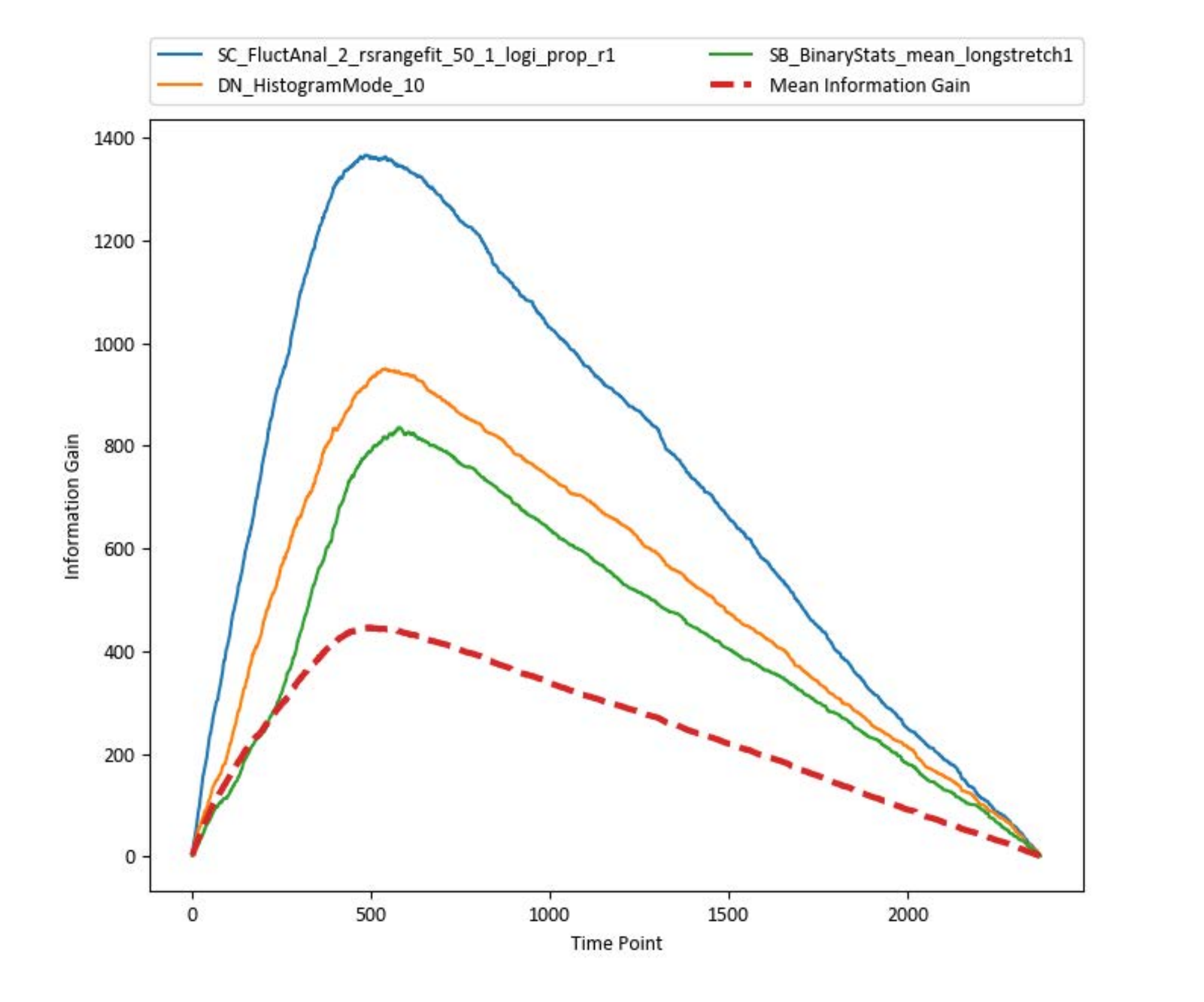}
            \caption{Top shows the example time series for the Asphalt-PavementType problem with class separated by colour. Bottom shows the temporal importance curves for the Asphalt-PavementType problem.}
            \label{asphaltTIC}
        \end{figure}
    	
	Figure~\ref{asphaltTIC} displays the CIF temporal importance curves for the Asphalt-PavementTypes problem with no subsampling.
    From this we can discern that time points in the range from 400-700 are the most important in determining pavement type.
    The most informative features in descending order of max information gain are mode of z-scored distribution using a 10-bin histogram, the longest period of consecutive values above the mean, and the proportion of slower timescale fluctuations that scale with linearly rescaled range fits. 
    
\section{Conclusion}
\label{conclusion}

Our contribution is to propose a new time series classifier, CIF, that combines the best elements of TSF and catch22 in a novel way. CIF is significantly more accurate than TSF and catch22. It is not significantly worse than the best other classifiers built on a single representation (WEASEL, S-BOSS, PF and STC) and is generally faster and requires less memory than them. When CIF replaces TSF in HIVE-COTE the resulting classifier, HC-CIF, is significantly more accurate than HIVE-COTE, TS-CHIEF and InceptionTime. HC-CIF represents a new state of the art for TSC in terms of classification accuracy on the UCR archive. 

There are many other improvements we could attempt with HIVE-COTE. For example, we could have used PF and S-BOSS instead of EE and BOSS. However, that would have introduced more sources of variation and obscured our core contribution: CIF is a new classifier that is best of its class, and by making the single change in HIVE-COTE of replacing TSF with CIF, the result is a classifier with significantly higher accuracy on average than the current state of the art.


\section*{Acknowledgements}{
     The experiments were carried out on the High Performance Computing Cluster supported by the Research and Specialist Computing Support service at the University of East Anglia. Our thanks to Carl Lubba for help with setting up and running the sktime catch22 version used in our initial investigations.
}

\bibliographystyle{ieee/IEEEtran}
\bibliography{paper}

\begin{thebibliography}{10}
\providecommand{\url}[1]{#1}
\csname url@samestyle\endcsname
\providecommand{\newblock}{\relax}
\providecommand{\bibinfo}[2]{#2}
\providecommand{\BIBentrySTDinterwordspacing}{\spaceskip=0pt\relax}
\providecommand{\BIBentryALTinterwordstretchfactor}{4}
\providecommand{\BIBentryALTinterwordspacing}{\spaceskip=\fontdimen2\font plus
\BIBentryALTinterwordstretchfactor\fontdimen3\font minus
  \fontdimen4\font\relax}
\providecommand{\BIBforeignlanguage}[2]{{%
\expandafter\ifx\csname l@#1\endcsname\relax
\typeout{** WARNING: IEEEtran.bst: No hyphenation pattern has been}%
\typeout{** loaded for the language `#1'. Using the pattern for}%
\typeout{** the default language instead.}%
\else
\language=\csname l@#1\endcsname
\fi
#2}}
\providecommand{\BIBdecl}{\relax}
\BIBdecl

\bibitem{bagnall2017great}
A.~Bagnall, J.~Lines, A.~Bostrom, J.~Large, and E.~Keogh, ``The great time
  series classification bake off: a review and experimental evaluation of
  recent algorithmic advances,'' \emph{Data Mining and Knowledge Discovery},
  vol.~31, no.~3, pp. 606--660, 2017.

\bibitem{lines2018time}
J.~Lines, S.~Taylor, and A.~Bagnall, ``Time series classification with
  hive-cote: The hierarchical vote collective of transformation-based
  ensembles,'' \emph{ACM Transactions on Knowledge Discovery from Data (TKDD)},
  vol.~12, no.~5, p.~52, 2018.

\bibitem{shifaz2020ts}
A.~Shifaz, C.~Pelletier, F.~Petitjean, and G.~I. Webb, ``Ts-chief: a scalable
  and accurate forest algorithm for time series classification,'' \emph{Data
  Mining and Knowledge Discovery}, pp. 1--34, 2020.

\bibitem{fawaz2019inceptiontime}
H.~I. Fawaz, B.~Lucas, G.~Forestier, C.~Pelletier, D.~F. Schmidt, J.~Weber,
  G.~I. Webb, L.~Idoumghar, P.-A. Muller, and F.~Petitjean, ``Inceptiontime:
  Finding alexnet for time series classification,'' \emph{arXiv preprint
  arXiv:1909.04939}, 2019.

\bibitem{deng2013time}
H.~Deng, G.~Runger, E.~Tuv, and M.~Vladimir, ``A time series forest for
  classification and feature extraction,'' \emph{Information Sciences}, vol.
  239, pp. 142--153, 2013.

\bibitem{lubba2019Catch22}
C.~H. Lubba, S.~S. Sethi, P.~Knaute, S.~R. Schultz, B.~D. Fulcher, and N.~S.
  Jones, ``catch22: Canonical time-series characteristics,'' \emph{Data Mining
  and Knowledge Discovery}, vol.~33, no.~6, pp. 1821--1852, 2019.

\bibitem{dau2019ucr}
H.~A. Dau, A.~Bagnall, K.~Kamgar, C.-C.~M. Yeh, Y.~Zhu, S.~Gharghabi, C.~A.
  Ratanamahatana, and E.~Keogh, ``The ucr time series archive,'' \emph{IEEE/CAA
  Journal of Automatica Sinica}, vol.~6, no.~6, pp. 1293--1305, 2019.

\bibitem{lucas2019proximity}
B.~Lucas, A.~Shifaz, C.~Pelletier, L.~O’Neill, N.~Zaidi, B.~Goethals,
  F.~Petitjean, and G.~I. Webb, ``Proximity forest: an effective and scalable
  distance-based classifier for time series,'' \emph{Data Mining and Knowledge
  Discovery}, vol.~33, no.~3, pp. 607--635, 2019.

\bibitem{lines2015time}
J.~Lines and A.~Bagnall, ``Time series classification with ensembles of elastic
  distance measures,'' \emph{Data Mining and Knowledge Discovery}, vol.~29,
  no.~3, pp. 565--592, 2015.

\bibitem{schafer2017fast}
P.~Sch{\"a}fer and U.~Leser, ``Fast and accurate time series classification
  with weasel,'' in \emph{Proceedings of the 2017 ACM on Conference on
  Information and Knowledge Management}, 2017, pp. 637--646.

\bibitem{large2019time}
J.~Large, A.~Bagnall, S.~Malinowski, and R.~Tavenard, ``On time series
  classification with dictionary-based classifiers,'' \emph{Intelligent Data
  Analysis}, vol.~23, no.~5, pp. 1073--1089, 2019.

\bibitem{bagnall2019chapter1}
\BIBentryALTinterwordspacing
A.~Bagnall, J.~Large, and M.~Middlehurst, ``A tale of two toolkits, report the
  second: bake off redux. chapter 1. dictionary based classifiers,''
  \emph{ArXiv e-prints}, vol. arXiv:1911.12008, 2019. [Online]. Available:
  \url{http://arxiv.org/abs/1911.12008}
\BIBentrySTDinterwordspacing

\bibitem{schafer2015boss}
P.~Sch{\"a}fer, ``The boss is concerned with time series classification in the
  presence of noise,'' \emph{Data Mining and Knowledge Discovery}, vol.~29,
  no.~6, pp. 1505--1530, 2015.

\bibitem{bostrom2017binary}
A.~Bostrom and A.~Bagnall, ``Binary shapelet transform for multiclass time
  series classification,'' in \emph{Transactions on Large-Scale Data-and
  Knowledge-Centered Systems XXXII}.\hskip 1em plus 0.5em minus 0.4em\relax
  Springer, 2017, pp. 24--46.

\bibitem{lines2012shapelet}
J.~Lines, L.~M. Davis, J.~Hills, and A.~Bagnall, ``A shapelet transform for
  time series classification,'' in \emph{Proceedings of the 18th ACM SIGKDD
  international conference on Knowledge discovery and data mining}.\hskip 1em
  plus 0.5em minus 0.4em\relax ACM, 2012, pp. 289--297.

\bibitem{flynn2019contract}
M.~Flynn, J.~Large, and A.~Bagnall, ``The contract random interval spectral
  ensemble (c-rise): the effect of contracting a classifier on accuracy,'' in
  \emph{International Conference on Hybrid Artificial Intelligence
  Systems}.\hskip 1em plus 0.5em minus 0.4em\relax Springer, 2019, pp.
  381--392.

\bibitem{baydogan2013bag}
M.~G. Baydogan, G.~Runger, and E.~Tuv, ``A bag-of-features framework to
  classify time series,'' \emph{IEEE transactions on pattern analysis and
  machine intelligence}, vol.~35, no.~11, pp. 2796--2802, 2013.

\bibitem{baydogan2016time}
M.~G. Baydogan and G.~Runger, ``Time series representation and similarity based
  on local autopatterns,'' \emph{Data Mining and Knowledge Discovery}, vol.~30,
  no.~2, pp. 476--509, 2016.

\bibitem{fawaz2019deep}
H.~I. Fawaz, G.~Forestier, J.~Weber, L.~Idoumghar, and P.-A. Muller, ``Deep
  learning for time series classification: a review,'' \emph{Data Mining and
  Knowledge Discovery}, vol.~33, no.~4, pp. 917--963, 2019.

\bibitem{fulcher2017hctsa}
B.~D. Fulcher and N.~S. Jones, ``hctsa: A computational framework for automated
  time-series phenotyping using massive feature extraction,'' \emph{Cell
  systems}, vol.~5, no.~5, pp. 527--531, 2017.

\bibitem{bagnall2018uea}
A.~Bagnall, H.~A. Dau, J.~Lines, M.~Flynn, J.~Large, A.~Bostrom, P.~Southam,
  and E.~Keogh, ``The uea multivariate time series classification archive,
  2018,'' \emph{arXiv preprint arXiv:1811.00075}, 2018.

\bibitem{demvsar2006statistical}
J.~Dem{\v{s}}ar, ``Statistical comparisons of classifiers over multiple data
  sets,'' \emph{Journal of Machine learning research}, vol.~7, no. Jan, pp.
  1--30, 2006.

\bibitem{garcia2008extension}
S.~Garcia and F.~Herrera, ``An extension on``statistical comparisons of
  classifiers over multiple data sets''for all pairwise comparisons,''
  \emph{Journal of Machine Learning Research}, vol.~9, no. Dec, pp. 2677--2694,
  2008.

\bibitem{shokoohi2017generalizing}
M.~Shokoohi-Yekta, B.~Hu, H.~Jin, J.~Wang, and E.~Keogh, ``Generalizing dtw to
  the multi-dimensional case requires an adaptive approach,'' \emph{Data mining
  and knowledge discovery}, vol.~31, no.~1, pp. 1--31, 2017.

\bibitem{karlsson2016generalized}
I.~Karlsson, P.~Papapetrou, and H.~Bostr{\"o}m, ``Generalized random shapelet
  forests,'' \emph{Data mining and knowledge discovery}, vol.~30, no.~5, pp.
  1053--1085, 2016.

\bibitem{middlehurst2019scalable}
M.~Middlehurst, W.~Vickers, and A.~Bagnall, ``Scalable dictionary classifiers
  for time series classification,'' in \emph{International Conference on
  Intelligent Data Engineering and Automated Learning}, ser. Lecture Notes in
  Computer Science.\hskip 1em plus 0.5em minus 0.4em\relax Springer, 2019, pp.
  11--19.

\bibitem{zhang20tapnet}
X.~Zhang, Y.~Gao, J.~Lin, and C.-T. Lu, ``{TapNet}: Multivariate time series
  classification with attentional prototypical network,'' in \emph{In proc.
  34th AAAI conference on artificial intelligence}, 2020.

\bibitem{schafer2017multivariate}
P.~Sch{\"a}fer and U.~Leser, ``Multivariate time series classification with
  weasel+ muse,'' \emph{arXiv preprint arXiv:1711.11343}, 2017.

\bibitem{souza2018asphalt}
V.~M. Souza, ``Asphalt pavement classification using smartphone accelerometer
  and complexity invariant distance,'' \emph{Engineering Applications of
  Artificial Intelligence}, vol.~74, pp. 198--211, 2018.

\bibitem{souza2017towards}
V.~M. Souza, E.~A. Cherman, R.~G. Rossi, and R.~A. Souza, ``Towards automatic
  evaluation of asphalt irregularity using smartphone’s sensors,'' in
  \emph{International Symposium on Intelligent Data Analysis}.\hskip 1em plus
  0.5em minus 0.4em\relax Springer, 2017, pp. 322--333.

\bibitem{batista2014cid}
G.~E. Batista, E.~J. Keogh, O.~M. Tataw, and V.~M. De~Souza, ``Cid: an
  efficient complexity-invariant distance for time series,'' \emph{Data Mining
  and Knowledge Discovery}, vol.~28, no.~3, pp. 634--669, 2014.

\end{thebibliography}

\end{document}